\documentclass[sigconf]{acmart}

\usepackage{wrapfig}

\AtBeginDocument{%
  \providecommand\BibTeX{{%
    \normalfont B\kern-0.5em{\scshape i\kern-0.25em b}\kern-0.8em\TeX}}}

\setcopyright{acmcopyright}
\copyrightyear{2022}
\acmYear{2022}
\setcopyright{acmcopyright}\acmConference[CIKM '22]{Proceedings of the 31st ACM International Conference on Information and Knowledge Management}{October 17--21, 2022}{Atlanta, GA, USA}
\acmBooktitle{Proceedings of the 31st ACM International Conference on Information and Knowledge Management (CIKM '22), October 17--21, 2022, Atlanta, GA, USA}
\acmPrice{15.00}
\acmDOI{10.1145/3511808.3557155}
\acmISBN{978-1-4503-9236-5/22/10}

\begin{document}

%\title{An Empirical Study on the Cumulative Effect of Multiple Fairness-Enhancing Interventions}
%\title{Debiasing Cascade: Studying the Cumulative Effect of Multiple Fairness-Enhancing Interventions}
\title{Cascaded Debiasing: Studying the Cumulative Effect of Multiple Fairness-Enhancing Interventions}

\author{Bhavya Ghai}
\affiliation{
    \institution{Stony Brook University} 
    \state{New York}
    \country{USA}}
\email{bghai@cs.stonybrook.edu}

\author{Mihir Mishra}
\affiliation{
    \institution{Los Altos High School} 
    \state{California}
    \country{USA}}
\email{mihir.mishra@gmail.com}

\author{Klaus Mueller}
\affiliation{
    \institution{Stony Brook University} 
    \state{New York}
    \country{USA}}
\email{mueller@cs.stonybrook.edu}

\renewcommand{\shortauthors}{Ghai, et al.}

\begin{abstract}
Understanding the cumulative effect of multiple fairness-enhancing interventions at different stages of the machine learning (ML) pipeline is a critical and underexplored facet of the fairness literature. Such knowledge can be valuable to data scientists/ML practitioners in designing fair ML pipelines. This paper takes the first step in exploring this area by undertaking an extensive empirical study comprising 60 combinations of interventions, 9 fairness metrics, 2 utility metrics (Accuracy and F1 Score) across 4 benchmark datasets. We quantitatively analyze the experimental data to measure the impact of multiple interventions on fairness, utility and population groups. We found that applying multiple interventions results in better fairness and lower utility than individual interventions on aggregate. However, adding more interventions do no always result in better fairness or worse utility. The likelihood of achieving high performance (F1 Score) along with high fairness increases with larger number of interventions. On the downside, we found that fairness-enhancing interventions can negatively impact different population groups, especially the privileged group. This study highlights the need for new fairness metrics that account for the impact on different population groups apart from just the disparity between groups. Lastly, we offer a list of combinations of interventions that perform best for different fairness and utility metrics to aid the design of fair ML pipelines.
\end{abstract}


%\ccsdesc[300]{Applied computing~Computers in other domains}


\begin{comment}
\begin{teaserfigure}
  \includegraphics[width=\textwidth]{sampleteaser}
  \caption{Seattle Mariners at Spring Training, 2010.}
  \Description{Enjoying the baseball game from the third-base
  seats. Ichiro Suzuki preparing to bat.}
  \label{fig:teaser}
\end{teaserfigure}
\end{comment}

\maketitle

\section{Introduction}
\begin{figure}
 \centering 
 \includegraphics[width=\columnwidth]{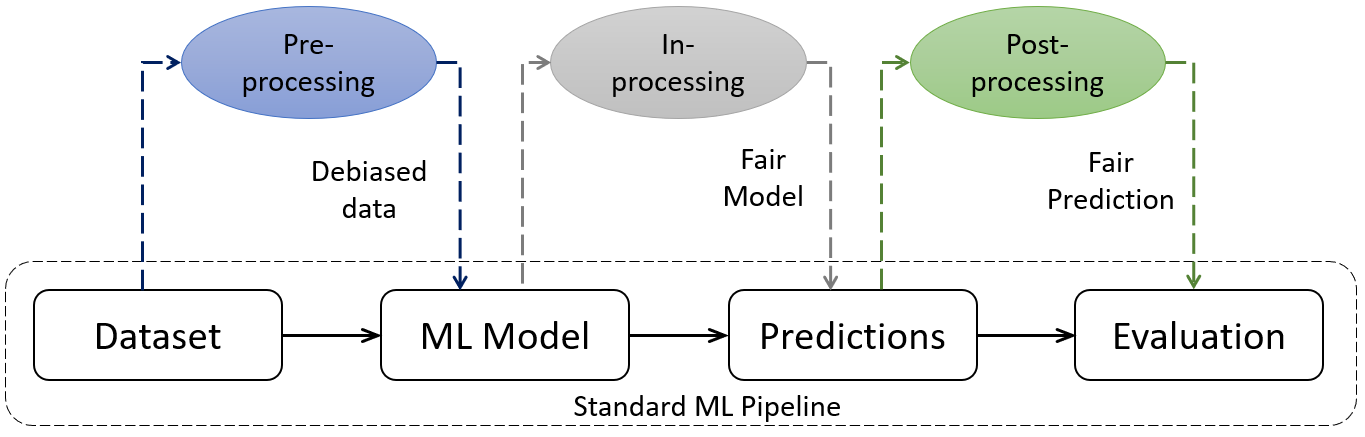}
 \vspace{-15pt}
 \caption{Three different types of fairness-enhancing interventions and how they fit into the standard ML pipeline.}  
 \label{fig:types}
\vspace{-1em}
\end{figure}

Algorithmic bias is a complex socio-technical problem whose impact can be felt in all sub-disciplines of machine learning \cite{100word,ghai2020measuring,vision,accent,ekstrand,ghai2021wordbias}. Recent years have seen a huge surge of fairness enhancing interventions that operate at different stages of the ML pipeline. 
%Some of these interventions are more effective than others at reducing bias as captured by a specific fairness metric. 
However, the problem is far from being solved 
%none of these interventions are able to curb the problem entirely 
if that is even possible \cite{friedler2021possibility}. Hence, there is a need for better interventions to reduce bias even further. Moreover, algorithmic bias can virtually emerge from any single or multiple stage(s) of the machine learning pipeline, right from problem formulation, dataset selection/creation to model formulation, deployment, and so on \cite{holstein2019improving}. The existing literature primarily focuses on curbing algorithmic bias by intervening at \textit{a} particular stage of the ML pipeline (see \autoref{fig:types}) \cite{hort2022bia}. However, algorithmic bias might still flourish via other stages/components of the ML pipeline. So, our focus should be on ensuring fairness across the ML pipeline instead of a single stage of the pipeline. This issue is also backed by a recent study with ML practitioners that elaborated on the disconnect between academic research and real world needs \cite{holstein2019improving}. One of the findings was to consider fairness as a system level property where the focus is on evaluating the impact of ML system as a whole instead of monitoring individual components.
 
%the net impact of machine learning systems is a culmination of all the individual contributions of the different stages of the ML pipeline. So,  
%Algorithmic bias is a complex socio-technical problem whose impact can be felt in all sub-disciplines of machine learning, be it Natural Language Processing \cite{100word, bios,boluk}, Recommendation Systems \cite{ekstrand, sweeney2013discrimination}, Speech \cite{accent}, Vision \cite{vision, convnets, genderShades}, Search \cite{ceo}, and many others.  

An intuitive solution to enhance fairness across the ML pipeline can be to apply multiple fixes (interventions) at different stages of the ML pipeline where bias can emerge from. We will refer to such a series of fairness-enhancing interventions as \textit{cascaded interventions}. For example, one might choose to debias the dataset, train a fairness-aware classifier over it and then post-process the model's predictions to achieve more fairness. 
%This approach is inline with the real world where different policies/guidelines/laws (interventions) are used to alleviate social inequality like Affirmative action in the US and Caste based reservation in India. Such interventions support the unprivileged group at different stages of life like education, employment, etc. 
This approach is inline with the real world where different laws/policies/guidelines try to alleviate social inequality by intervening at multiple stages of life like education, employment, promotion, etc. Examples include Affirmative action in the US and Caste-based reservation in India. 
%For example, the Indian Government provides caste based reservation (policy intervention) to support the underprivileged groups (Scheduled Castes, Scheduled Tribes and Other Backward Classes) get an easier access to educational institutions, jobs, etc. . 
%In the AI world, such an approach  
This begs the question if it were possible to achieve more fairness in the ML world by intervening at multiple different stages of the ML pipeline
%applying a series of interventions 
and what might be its possible fallouts. 
%This question is especially pertinent to industry practitioners who develop ML systems that directly impact the lives of people.  In the rest of the paper, we will refer to fairness enhancing interventions in the AI world as just interventions and a series of fairness enhancing interventions as cascaded interventions. 
   
%There has been a surge in fairness enhancing interventions across different stages of the ML pipeline. Perfect fairness, if it exists, might be impossible to achieve as different fairness metrics can be mutually incompatible. Existing interventions focus on a specific facet of fairness and claim to reduce bias based on the corresponding fairness metric. However, such interventions are evaluated without much regard for the possible fallout on other fairness metrics and the privileged group. In this work, we

%Existing literature on Algorithmic Fairness focuses heavily on novel individual interventions and not so much attention has been paid towards the net impact of cascaded multiple interventions. Similar to the real world, it is important to understand the possible implications of such an intervention in the AI world before deployment. 
%Impact of algorithms on the real world is due to the entire ML pipeline and not just due to a particular stage in the ML pipeline. So, we focus on building fail ml pipelines and not just fair components. 
%Can we use different interventions as different components of the fair ML pipeline. 

\begin{figure*}
 \centering 
 \includegraphics[width=1.9\columnwidth]{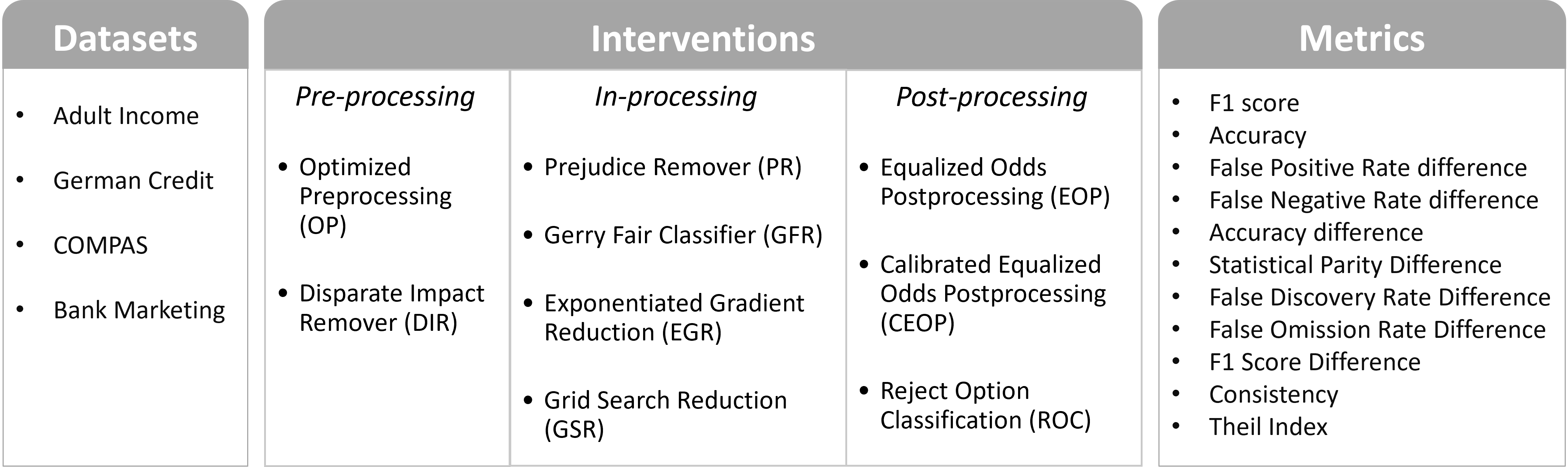}
 \vspace{-5pt}
 \caption{Experimental Setup - Different datasets, interventions and metrics considered for the empirical study}  
 \label{fig:exp_setup}
\end{figure*}

In this work, we undertake an extensive empirical study to understand the impact of individual interventions as well as the cumulative impact of cascaded interventions on utility metrics like accuracy, different fairness metrics and on the privileged/unprivileged groups. Here, we have focused on the binary classification problem over tabular datasets with a single sensitive (protected) attribute. We have considered 9 different interventions where 2 operate at the data stage, 4 operate at the modeling stage and 3 operate at the post-modeling stage. We also consider all possible combinations of these interventions as shown in \autoref{fig:distribution}. To execute multiple interventions in conjunction, we feed the output of one intervention as input to the next stage of the ML pipeline. We simulate multiple three-stage ML pipelines that are acted upon by different combinations of interventions. 
%For example, one might choose to debias the dataset, train a fairness aware classifier over it and then post-process the model's predictions to achieve more fairness. 
We measure the impact of all these interventions on 9 fairness metrics and 2 utility metrics over 4 different datasets. Thereafter, we perform quantitative analyses on the results and try to answer the following research questions:

\begin{itemize}
    \item [\textbf{R1.}] Effect of cascaded interventions on fairness metrics \\ Does intervening at multiple stages reduce bias even further? If so, does it always hold true? 
    %What is the aggregate relationship between different number of interventions in terms of fairness?
    %How do different numbers of interventions compare in terms of fairness? 
    What is the impact on group fairness metrics and individual fairness metrics?
    \item [\textbf{R2.}] Effect of cascaded interventions on utility metrics \\
    How do utility metrics like accuracy and F1 score vary with different number of interventions? Existing literature discusses the presence of fairness-utility tradeoff for individual interventions. Does it hold true for cascaded interventions?  
    \item [\textbf{R3.}] Effect of cascaded interventions on population groups \\
    How are the privileged and unprivileged groups impacted by cascaded interventions in terms of F1 score, false negative rate, etc.? Are there any negative impacts on either groups?
    
    \item [\textbf{R4.}] How do different cascaded interventions compare on fairness and utility metrics?
    %Do certain combinations of debiasing techniques work better than others?
\end{itemize}

%The source code for this work is attached to the supplementary material and will be made publicly available upon acceptance for easy reproducibility. 

%\textcolor{blue}{
%- Need for multiple interventions (find references). Lack of any study which investigates the impact \\
%- Our contribution
%}

%The inspiration for this work originates in the several accusations of reverse discrimination that have emerged in recent years. We may ask, in trying to counter fairness, do algorithmic discriminate against the privileged class?

\section{Background and Literature Review}

\subsection{Fairness-Enhancing Interventions}
\label{sec:interventions}
Bias mitigation techniques can be broadly classified into 3 stages :- Pre-processing, In-processing and Post-processing (Fig. \ref{fig:types}). In the following, we discuss a few interventions that we have considered in this work, in the context of the intervention stage they operate. 

\subsubsection{Pre-processing} Interventions at the Pre-processing stage operate on the raw dataset to generate its debiased version. The debiased dataset can then be fed back into the standard ML pipeline for fairer predictions. Specifically: 

%\emph{Reweighing (RW) —  Preprocessing technique applying weights on group/label combinations to achieve fair results.} \cite{kamiran2012data}

\emph{Optimized Preprocessing (OP)} —  uses convex optimization to transform the underlying dataset such that fairness is enhanced and utility is preserved with limited data distortion \cite{calmon2017optimized}.

\emph{Disparate Impact Remover (DIR)} — edits the feature set of a given dataset such that the predictability of the protected variable is impossible % distinguish between groups is curbed 
%to achieve similar base rates for the privileged and underprivileged groups 
while preserving rank ordering within groups \cite{feldman2015certifying}.

\subsubsection{In-processing} Interventions in this stage operate at the data modeling stage to train a fair ML model. Specifically: 

\emph{Gerry Fair Classifier (GFC)} — formulates fairness as a zero-sum game between a Learner (the primal player) and an Auditor (the dual player) to compute an equilibrium for this game \cite{kearns2018preventing}. 
%game where \textcolor{blue}{algorithm for learning classifiers that are fair with respect to rich subgroups} .

\emph{Prejudice Remover (PR)} — adds a specialized regularization term to the learning objective
such that the classifier becomes independent of the sensitive information
%to train a discrimination-aware ML model 
\cite{kamishima2012fairness}.

\emph{Exponential Gradient Reduction (EGR)} — reduces fair classification to a sequence of cost-sensitive classification problems, returning a \textit{randomized classifier} with the lowest empirical error subject to fair classification constraints \cite{agarwal2018reductions}.

\emph{Grid Search Reduction (GSR)} — reduces fair classification to a sequence of cost-sensitive classification problems, returning the \textit{deterministic classifier} with the lowest empirical error subject to fair classification constraints \cite{agarwal2018reductions, agarwal2019fair}.

%\emph{Gerry Fair Classifier (GFC)}
%\emph{Meta Fair Classifier}

\subsubsection{Post-processing} Such interventions operate on the model's predictions to yield more fair predictions. Specifically: 

\emph{Calibrated Equalized Odds Postprocessing (CEOP)} — changes classifier results based on calibrated score outputs and an equalized odds goal \cite{pleiss2017fairness}.

\emph{Equalized Odds Postprocessing (EOP)} — solves a linear program to find probabilities whose corresponding labels will optimize the equalized odds goal \cite{hardt2016equality, pleiss2017fairness}.

\emph{Reject Option Classification (ROC)} — reduces discrimination by assigning positive labels to the unprivileged groups and negative labels to the privileged groups for the data points that lie close to the decision boundary, i.e., the points for which the classifier is uncertain about \cite{kamiran2012decision}.

Existing literature has studied the above mentioned interventions in isolation. In this work, we explore if a combination of these interventions can lead to enhanced fairness across the ML pipeline. 
%lead mostly focused on a single intervention technique. In this paper, we study the impact of multiple interventions on different fairness and utility metrics. 

\begin{figure}
    \centering
    \includegraphics[scale=0.45]{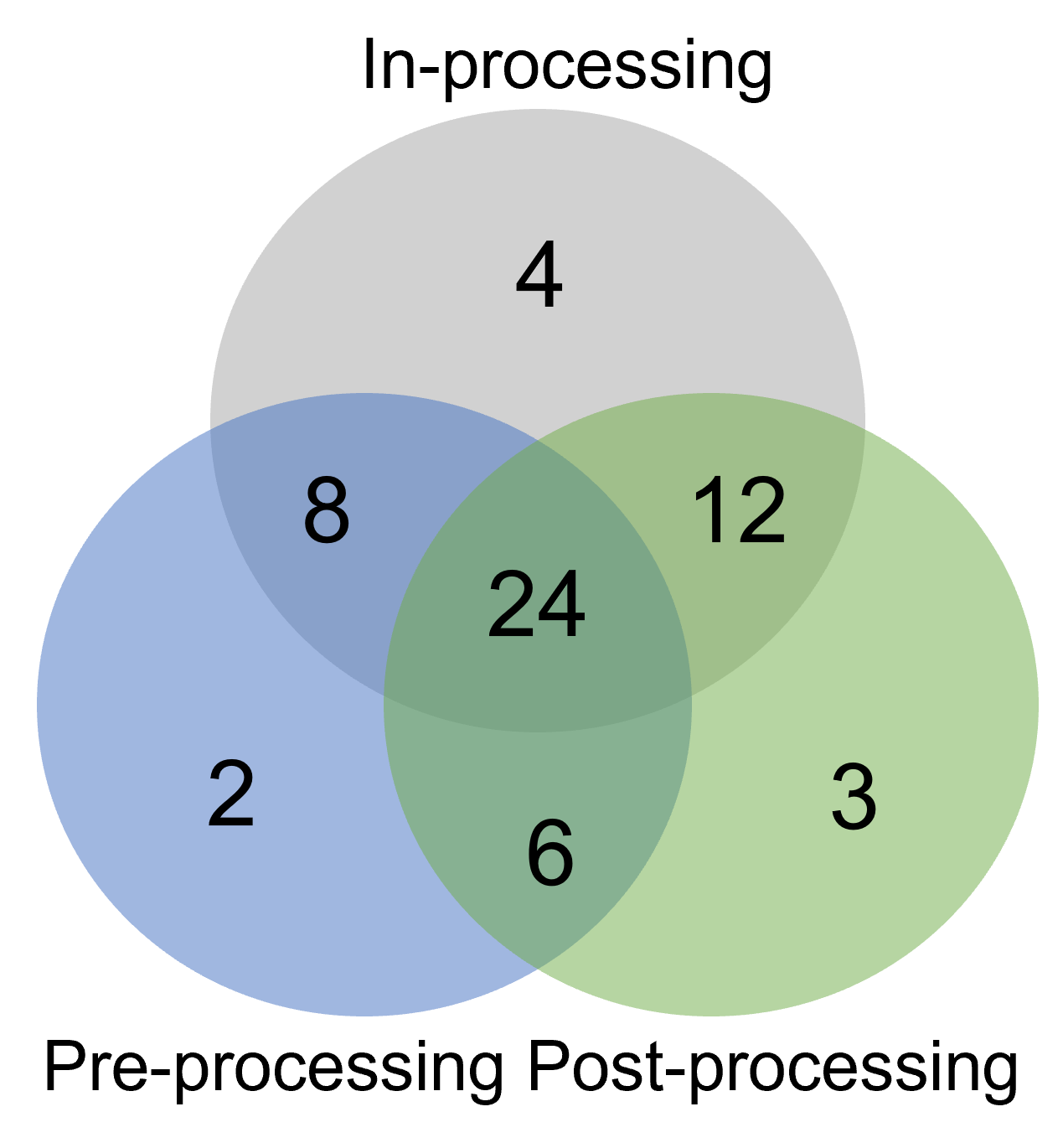}
    \caption{Distribution of fairness-enhancing interventions considered in this paper. This includes 9 individual interventions and 50 different combinations of interventions.}  
    \label{fig:distribution}
\end{figure}

\subsection{Measuring Fairness}
Quantifying fairness for ML algorithms is an active research area. Numerous fairness metrics have been proposed in the literature which mathematically encode different facets of fairness like group fairness, individual fairness, counterfactual fairness, etc. \cite{dwork2012fairness,dwork2018group,aif360, bechavod2017penalizing, kate,kusner2017counterfactual,arvindTalk}. For e.g., group fairness implies that members of one group should receive a similar proportion of positive/negative outcomes as other groups \cite{bechavod2017penalizing, kate}, individual fairness implies that similar individuals should be treated similarly \cite{dwork2012fairness, dwork2018group}, etc. 
%%%Another way to classify fairness metrics can be on the level they operate on. For e.g., dataset-based metrics are solely calculated on the basis of the dataset and are independent of the classifier. On the other hand, classifier-based metrics are dependent on the predictions of the classifier like the false negative rate difference. 
%There are multiple fairness metrics which might be incompatible with each other. 
In this work, we have opted for a diverse set of 9 fairness metrics to paint a more comprehensive picture. 
%%%Here, we have not used any dataset based metrics due to their inability to capture the impact of in-processing and post-processing interventions.  

%%%It is important to note that fairness metrics do not capture the impact on different population groups. For example, the fairness metric, False Negative Rate difference, reports the difference in false negativity rate between groups. An increase/decrease in this metric does not tell us anything about the specific impact on the privileged or unprivileged groups. So, in this work, we also measure the specific impact on different population groups.
The efficacy of fairness-enhancing interventions is typically measured using different fairness metrics. However, these metrics do not directly capture the impact on different population groups. %These fairness metrics provide elegant mathematical representations for different notions of fairness but might obscure the effect on groups. 
For example, the impact on different population groups is irrelevant for fairness metrics following the notion of individual fairness. This even holds true for fairness metrics based on the notion of group fairness. Such metrics focus on measuring the disparity between groups without much regard for the impact on specific groups. For example, the fairness metric, False Negative Rate difference, reports the difference in false negative rate between groups. 
%and abstracts the impact on the actual groups. 
An increase/decrease in this metric does not tell us anything about the specific impact on the privileged or unprivileged groups. In this work, we analyze how interventions impact different fairness metrics and population groups.   
%It is possible that a reduction in False negativity rate difference might be due to an      

%Existing literature mostly focuses on the impact of ML algorithms and the different fairness enhancing interventions from the perspective of the underprivileged group.   
%There are plenty of studies which shows that Algorithms can discriminate against the minority class. However, in this paper, we are trying to explore the opposite. For instance. studies have shown that ML systems like word embedding models might associate privileged groups with positive stereotypes \cite{ghai2021wordbias}.  

%\subsection{Accuracy Fairness Trade-Off}
%\textcolor{blue}{It has been shown for individual interventions but what happens for multiple interventions is still unknown.} 

\subsection{Fairness across ML Pipeline}
Research that focuses on fairness for a multi-stage ML system has received some attention and is still in its early stages \cite{martin2021engineering, biswas2021fair, wang2021practical, wu2021fair, hort2022bia}. Biswas et al. studied the impact of data preprocessing techniques like standardization, feature selection, etc. on the overall fairness of the ML pipeline \cite{biswas2021fair}. They found certain data transformations like sampling to enhance bias. Hirzel et al. also focus on the data preprocessing stage \cite{martin2021engineering}. They present a novel technique to split datasets into train/test that is geared towards fairness. Wang et al. focused on fairness in the context of multi-component recommender systems \cite{wang2021practical}. They found that overall system’s fairness can be enhanced by improving fairness for individual components. 
%It is important to note that there is a related line of work that talks about fairness in compound decision making processes (cohort pipelines) \cite{bower2017fair}. Cohort pipelines differ from ML pipelines as the number of data points (individuals) gets reduced at every stage. They deal with
Our work focuses on how different combinations of interventions at 3 stages of the ML pipeline can be leveraged to enhance fairness across the ML pipeline.
%In this work, we focus on 3 stage ML pipelines that undergo different combinations of interventions. Our focus is to understand the net impact of a series of interventions. 

There is a related line of work that discusses fairness in the context of compound decision making processes \cite{bower2017fair, dwork2020individual, emelianov2019price}. In such data systems, there is a sequence of decisions to be made where each decision can be thought of as a classification problem. 
%For every decision, there is a pool of candidates who gets filtered and the remaining candidates proceed to the next stage. 
For eg., in a two stage hiring process, candidates are first filtered for the interview stage and the remaining candidates are again filtered to determine who gets hired. This line of work focuses on fairness over multiple tasks and does not pay much attention towards enhancing fairness of an individual task. 
%where each task (decision) can be modeled as a classification problem. Here, they don't pay much attention 
%For a two stage hiring process, Bower et al. showed that composition of fair components does not guarantee a fair pipeline as measured by a single fairness metric (equal opportunity). 
%Here, each decision can be  
%Cohort pipelines differ from ML pipelines as the number of data points (individuals) gets reduced at every stage. \textcolor{blue}{They deal with}  . 
 %and does not pay much attention towards ensuring fairness where each task can be a ML pipeline in itself. 
 Here, a single task (classification problem) can be thought of as a ML pipeline. This is where our work comes in. 
 %Our work is focused on enhancing fairness for a single task.  
 Our work studies the different combinations of interventions that can together enhance fairness for a single decision making process.

%Bower et al. highlighted the need to study fairness in the context of compound decision making processes . For a two stage hiring process, they showed that composition of fair components does not guarantee a fair pipeline as measured by a single fairness metric (equal opportunity).
%Fair AutoML \cite{wu2021fair}.

%\subsection{Impact on Population Groups}

\section{Experiment Setup}
We have used IBM's AIF 360 \cite{bellamy2019ai} open source toolkit to conduct all experiments for this paper. More specifically, we leveraged 4 datasets, 9 fairness enhancing interventions and 11 evaluation metrics from this toolkit as shown in Fig. \ref{fig:exp_setup}. To have a more even comparison, we have used the same ML model, i.e., logistic regression (linear model) across the board. Moreover, we only selected those in-processing interventions that are based on or compatible with linear models. 
%We only selected interventions which have . 
%So, we ignored interventions like Meta fair classifier. 

\begin{figure*}
 \centering 
 \includegraphics[width=1.4\columnwidth]{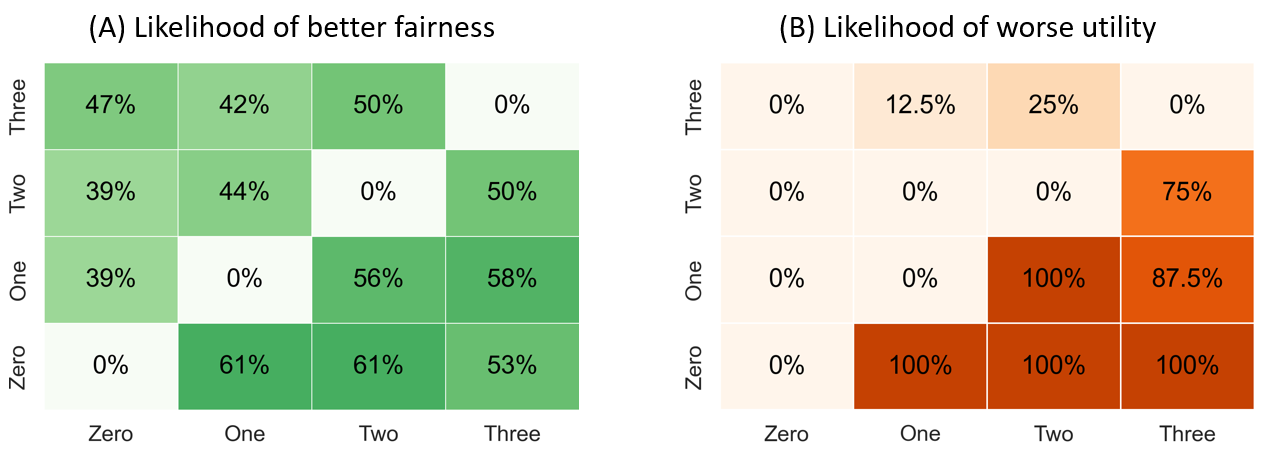}
 \caption{Heatmaps for Fairness and Utility metrics across different numbers of interventions. In Figure (A), a cell (i,j) represents the percentage of cases where j interventions yielded better fairness metrics than i interventions. In Figure (B), a cell (i,j) represents the percentage of cases where j interventions yielded worse utility metrics than i interventions. Here, i represents rows and j represents columns.
 %\textcolor{blue}{explain what direction i and j are}
 %In figure (B), a cell (i,j) represents the percentage of cases where i interventions yielded better utility metrics than j interventions. 
 }  
 \label{fig:heatmap}
\end{figure*}

\paragraph{\textbf{Interventions}} Among the 9 interventions, 2 belong to the pre-processing stage, 4 belong to the in-processing stage and 3 belong to the post-processing stage. Apart from these individual interventions, we also execute different combinations of these interventions in groups of 2 and 3. For example, one might choose to intervene at any 2 stages (say a pre-processing intervention followed by a post-processing intervention) or choose to intervene at all 3 stages of the ML pipeline. To form all possible combinations, we cycle through all available options (interventions) for a given ML stage along with a `No Intervention' option and repeat it for all the 3 stages.   
%We have conducted all possible combinations of interventions based on the 8 individual interventions that we have considered. 
This results in 8 combinations of pre-processing and in-processing interventions, 12 combinations of in-processing and post-processing interventions, 6 combinations of pre-processing and post-processing interventions and 24 combinations of of all 3 types of interventions (see Fig. \ref{fig:distribution}). In totality, we perform 9 individual interventions, 50 different combinations of interventions and a baseline case (No intervention for all stages) for each of the 4 datasets. Here, we have used the default set of hyperparameters for all interventions. In this paper, we will refer to the different interventions by their acronyms like PR for Prejudice Remover as defined in \autoref{sec:interventions}. For cascaded interventions, we will concatenate the respective acronyms with a `+' sign. For example, OP + PR means that we performed the Optimized Preprocessing (OP) intervention followed by the Prejudice Remover (PR) intervention. The baseline case is referred as `Logistic Regression'.

\paragraph{\textbf{Evaluation Metrics}} The impact of the different interventions is captured using a diverse set of 11 evaluation metrics. Two of them, namely accuracy and F1 score, are utility metrics that measure the ability of a ML model to learn the underlying patterns from the training dataset. Here, we have included the F1 score as it can better deal with imbalanced output class distributions. Both of these metrics range between 0 and 1. Higher values mean better performance. The remaining 9 metrics each capture some facet of fairness. Two of the fairness metrics, namely Consistency and Theil index, subscribe to the notion of individual fairness. Here, the Consistency metric can be understood as the degree to which k-nearest neighbors for different instances having the same output labels (k=5). 
%is computed using k-nearest neighbors algorithm to measure the degree to which similar instances have simlar output labels
On the other hand, Theil index comes from the family of inequality measures called the Generalized Entropy Measures \cite{aif360}. Higher values for Consistency and lower values for the Theil index mean more fairness. All other fairness metrics subscribe to the notion of group fairness, namely false positive rate difference (FPR Diff), false negative rate difference (FNR Diff), statistical parity difference (SPD), false discovery rate difference (FDR Diff), false omission rate difference (FOR Diff), accuracy difference (Accuracy Diff) and F1 score difference (F1 Score Diff). All group fairness metrics measure disparity between groups based on some measure such as false positive rate (FPR). A lower absolute value for the group fairness metrics means more fairness. The sign of these metrics represents the group that is getting the upper/lower hand. A value of 0 means perfect fairness.

\paragraph{\textbf{Datasets}} Each of the 4 tabular datasets used in this paper, as listed in \autoref{fig:exp_setup}, have been used extensively in the fairness literature. They deal with a binary classification problem and typically contain one or more binary protected attributes such as gender, race, etc. 
%%%For each of these datasets, we have used the default preprocessing procedure as provided by the AIF360 package (not to be confused with preprocessing interventions). It should be noted that the default preprocessing often involves one hot encoding to deal with categorical variables; this inflates the number of columns compared to the original dataset. 
We describe the datasets briefly as follows: 
%\begin{comment}
\paragraph{Adult Income Dataset} After pre-processing, this dataset consists of 45,222 rows and 99 columns that are derived from the 1994 Census database. Each row represents a person characterized by %a mix of numerical and categorical 
variables like education, gender, race, workclass, etc. These attributes are used to predict if an individual makes more than \$50k a year. Here, we have used gender as the sensitive attribute with males as the privileged group and females as the unprivileged group.  

\paragraph{German Credit Dataset} After pre-processing, this dataset consists of 1,000 rows and 59 columns which was originally prepared by Prof. Hofmann. The task is to predict if an individual has good or bad credit risk based on features like credit amount, credit history, sex, etc. Here, the sensitive attribute is age. Individuals older than 25 years belong to the privileged group and vice versa. 

\paragraph{COMPAS Recidivism Dataset} After pre-processing, this dataset contains 6,167 rows and 402 columns which pertains to the COMPAS algorithm used for scoring defendants in Broward County, Florida. The task is to predict if an individual will recommit a crime within a two year period based on personal attributes like charge degree, prior count, etc. Here, the sensitive attribute is race with Caucasians as the privileged group and vice versa.
%non-Caucasians as the unprivileged group. 

\paragraph{Bank Marketing Dataset} After pre-processing, this dataset consists of 30,488 rows and 53 columns; it pertains to a direct marketing campaign of a Portuguese banking institution. The classification task is to predict if a client will buy a term deposit based on features like type of job, marital status, education, etc. Here, the sensitive attribute is age. Individuals (clients) younger than 25 years belong to the unprivileged group and vice versa. 
%is binned into a binary categorical variable. \\
%\end{comment}
\paragraph{\normalfont {The positive outcome label for all these datasets refer to the favorable outcome for the recipient. For e.g., the positive outcome label for the adult income dataset refers to an income greater than \$50k. Similarly, for the COMPAS dataset, it refers to \textit{not} recommitting a crime in 2 years. This information will help interpret measures like false positive rate, false negative rate, etc. 
%For more details on the datasets, please refer to appendix A.
}}

\begin{figure*}[t]
 \centering 
 \includegraphics[width=1.5\columnwidth]{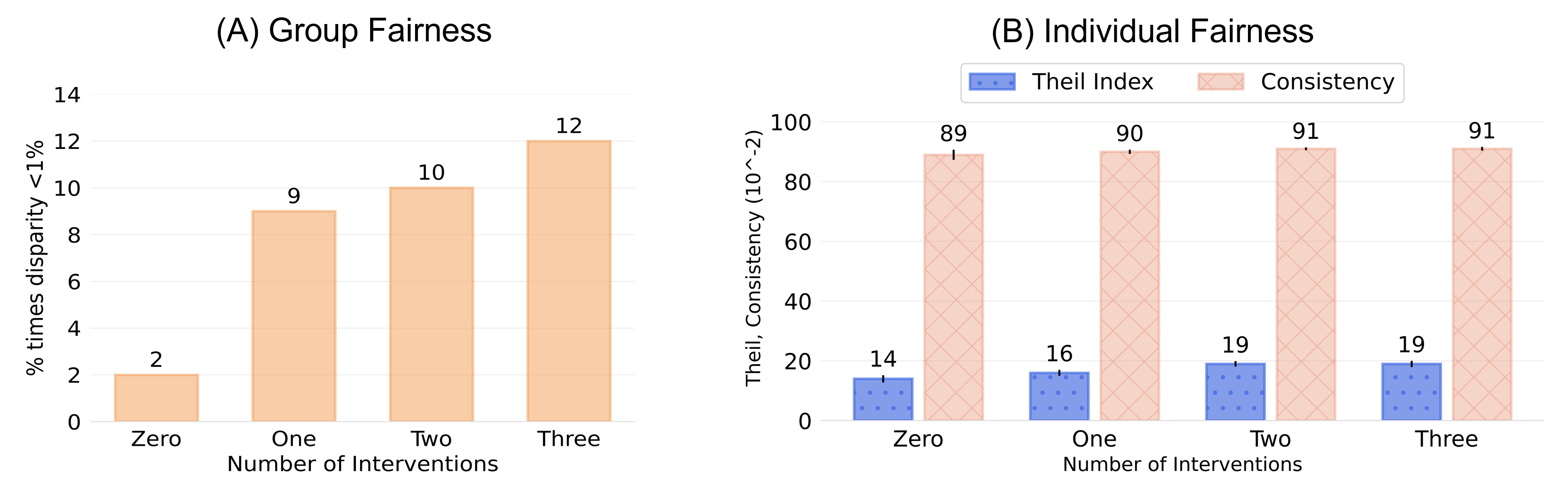}
  \setlength{\belowcaptionskip}{-2pt}
  \setlength{\abovecaptionskip}{-1pt}
 \caption{Effect of individual and cascaded interventions on fairness metrics (A) Percentage of times the disparity between the privileged and unprivileged groups was less than 1\% across all group fairness metrics. A higher value means more group fairness. (B) Mean values for Theil index and Consistency across different numbers of interventions. Lower values for Theil index and higher values for Consistency means more individual fairness. Error bars represent standard error.}  
 \label{fig:group_indi_inter}
\end{figure*}

\paragraph{\textbf{Method}} 
%%%After default pre-processing, we standardize different features of the dataset so that all non-protected features have the same mean and standard deviation. Thereafter, 
Each dataset is randomly divided into train and test dataset in the ratio 70:30. For the baseline case, we train a logistic regression model on the training dataset and then compute different evaluation metrics using the test data. Next, we execute all individual and cascaded interventions using the train dataset and record their impact on different utility and fairness metrics using the test dataset. Apart from these metrics, we also record statistics like false negative rate (FNR), base rate, etc. for the privileged and unprivileged groups.    
%Here, we have used two utility metrics i.e., Accuracy and F1 score and 8 fairness metrics. The fairness metrics can be bifurcated into 2 individual fairness metrics (theil index and consistency) and 6 group fairness metrics (see \autoref{fig:exp_setup}). 
This entire process is repeated 3 times for each dataset with different random splits between train and test dataset to counter sampling bias. Lastly, we compute the mean values for all evaluation metrics across the 3 iterations for each intervention. For each dataset, these results can be represented in tabular format with 60 rows and 11 columns where each row represents an intervention and each column represents an evaluation metric. 

% https://archive.ics.uci.edu/ml/datasets/bank+marketing

%Datasets - \textcolor{blue}{briefly explain different datasets , number of rows, features, output variable, sensitive features}

%\vspace{-1em}

\section{Results}
In this section, we analyze the empirical data from our experiments to understand the effect of different individual and cascaded interventions on fairness, utility and population groups.
%We analyze the empirical data from our experiments to study the effect of cascaded interventions on fairness, utility and population groups.
%answer all the research questions laid in the Introduction section based on our empirical data. 

%\subsection{Do Multiple Interventions reduce bias even further?}
%\subsection{Effect of Cascaded Interventions on Fairness Metrics}
\subsection{Effect on Fairness Metrics (R1)}
\label{sec:res_fair}

We first gauge the effect of cascaded interventions on fairness as a whole (across fairness metrics). We start by grouping all interventions into 4 buckets, i.e., 0 intervention, 1 intervention, 2 interventions and 3 interventions, respectively. For each bucket, we compute the average score for different fairness metrics and repeat this process for all datasets. It is important to note that different fairness metrics are not directly comparable as they are based on different interpretations of fairness and also vary in terms of their numerical distribution (range, mean, standard deviation, etc.). So, we will compare the mean value of a fairness metric with its counterpart for a different bucket. 
%Next, we compare the mean value for one bucket with another and count the percentage of times one bucket performs better than another.
We count the percentage of times one bucket performs better than another across fairness metrics and datasets. This data is visualized using a heatmap of size 4 x 4 in \autoref{fig:heatmap}(A). Each row and column represents a bucket (number of interventions). Here, a cell (i,j) represents the percentage of times j interventions performs better than i interventions. For example, the cell (2,1) is labeled 44\%. It means that a single intervention yielded better fairness scores than two interventions for 44\% of cases. A bucket j will be considered favorable over another bucket i if the value for the cell (i,j) is greater than 50\% and vice versa. 

\begin{figure}
 \centering 
 \includegraphics[width=1\columnwidth]{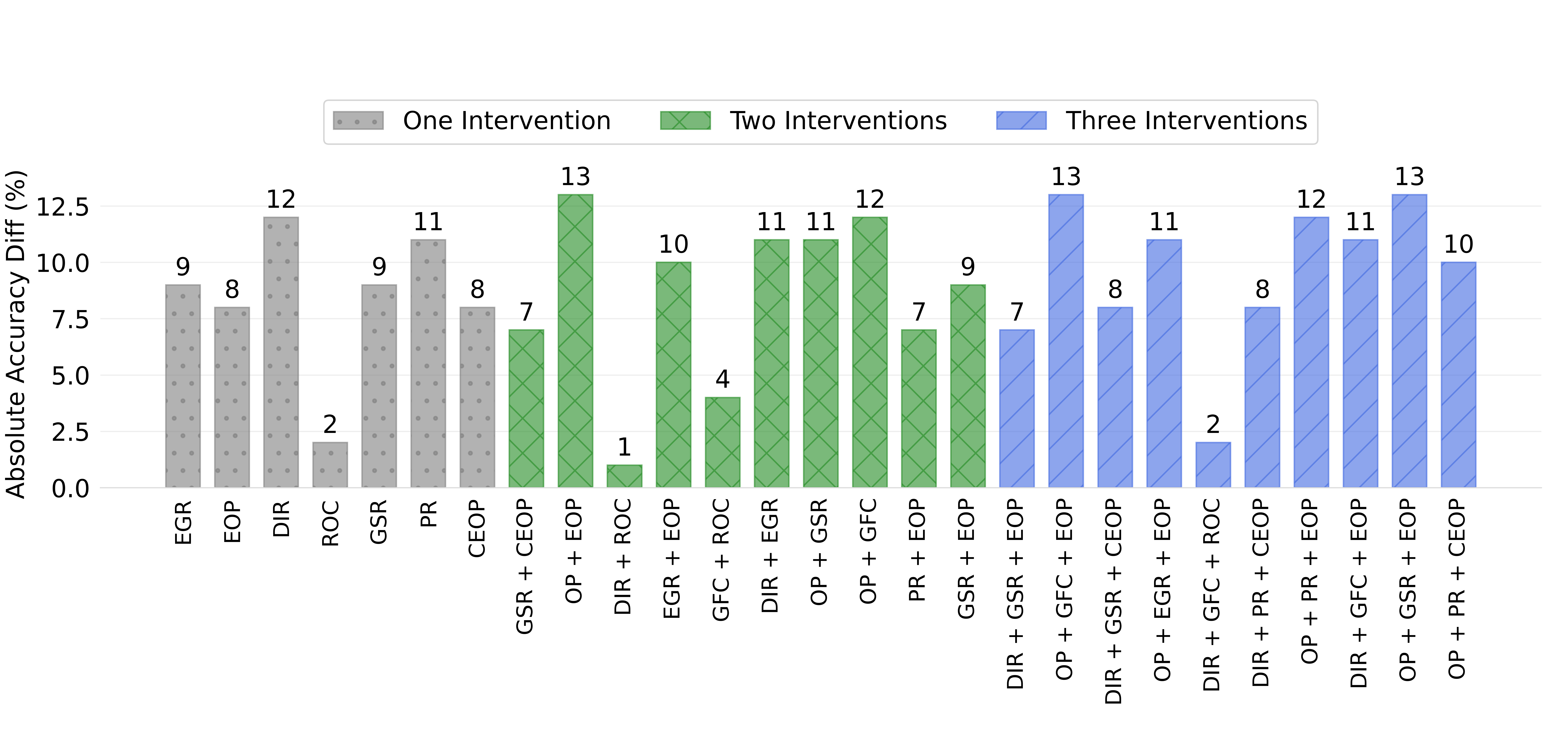}
  \setlength{\belowcaptionskip}{-8pt}
  \setlength{\abovecaptionskip}{-4pt}
%%% \caption{Absolute values for the Accuracy difference metric across different interventions for the Adult Income dataset. Here, lower values are desirable. This plot shows multiple cases where values corresponding to more number of interventions are larger than lower number of interventions. Hence, more interventions does not always lead to more fairness.}  
\caption{Absolute values for the Accuracy difference metric across different interventions for the Adult Income dataset. This plot shows multiple cases where more interventions does not always lead to more fairness.}
 \label{fig:counter_exp}
\end{figure}

It should be noted that different fairness metrics might be incompatible with each other. So, the net trend (cell values) might appear a bit faded as some fairness metrics might cancel the effect of another. Looking at the row i=0, we find that any number of interventions greater than 0 provide better overall fairness than having no interventions. Looking at the row i=1, we find that the columns j=2 and j=3 have values more than 50\%, i.e., two or three interventions yielded better fairness than a single intervention. Moving to the row i=2, we find that the value of the cell (2,3) is 50\%. Perhaps surprisingly, this means that it is equally likely for either buckets to outperform each other. Overall, it appears that fairness improves from 0 to 2 interventions and becomes constant thereafter. However, it is important to note that the heatmap encodes frequency and not the magnitude of difference between fairness metrics. So, it is possible that three interventions reduce bias significantly more (in terms of magnitude and not the count of fairness metrics) than the two interventions case and might still appear to be no better than the two interventions case. 
%cells with the same value might focus on the same fairness metrics but the extent to which they reduce bias might vary significantly.
%\textcolor{blue}{anything you can observe that is surprising or important? Why would I care about this result? Is it that better than 50\% is good and less is bad? As in 50\% is just chance? It should be stated. Would it be better to present it as odds? When I look at these cells none of these greater 50 numbers are really high. I think this needs more explanation of the significance of this result.}       

%A heatmap of size 4 x 4 where each cell represents the percentage of cases where the bias score reduces when comparing between two intervention groups. Here, each row and column represents an intervention group i.e., no intervention, 1 intervention, 2 interventions and 3 interventions respectively. For example, going from ‘no intervention’ (row) to ‘3 intervention’ (column) might have a value of 80%. This means that 80% of the times we observed a reduction of different bias metrics for interventions corresponding to ‘3 intervention’ group compared to no intervention group.  

The heatmap provides an aggregate picture of how fairness metrics vary for different numbers of interventions. Now, let us dig a bit deeper and gauge the impact of cascaded interventions on group fairness and individual fairness. For individual fairness, we plot the mean values for the Theil index and Consistency for different numbers of interventions. For the group fairness metrics, we compute the percentage of times the absolute value of each constituting metric is less than 0.01. As we can see from \autoref{fig:group_indi_inter} (A), the percent of times the group fairness metrics are below a threshold increases steadily with higher numbers of interventions from 2\% to 12\%. In other words, \textit{group fairness improves with more interventions on aggregate.} This observation largely concurs with our findings from \autoref{fig:heatmap}(A). On the other hand, we get mixed signals from the individual fairness metrics. The Consistency metric shows a slight improvement in fairness while the Theil index shows a downfall in fairness with higher numbers of interventions.
%(lower values for the Theil index are desirable). 

\begin{figure*}
    \centering
    \begin{minipage}{.33\textwidth}
        \centering
        \includegraphics[width=0.9\textwidth]{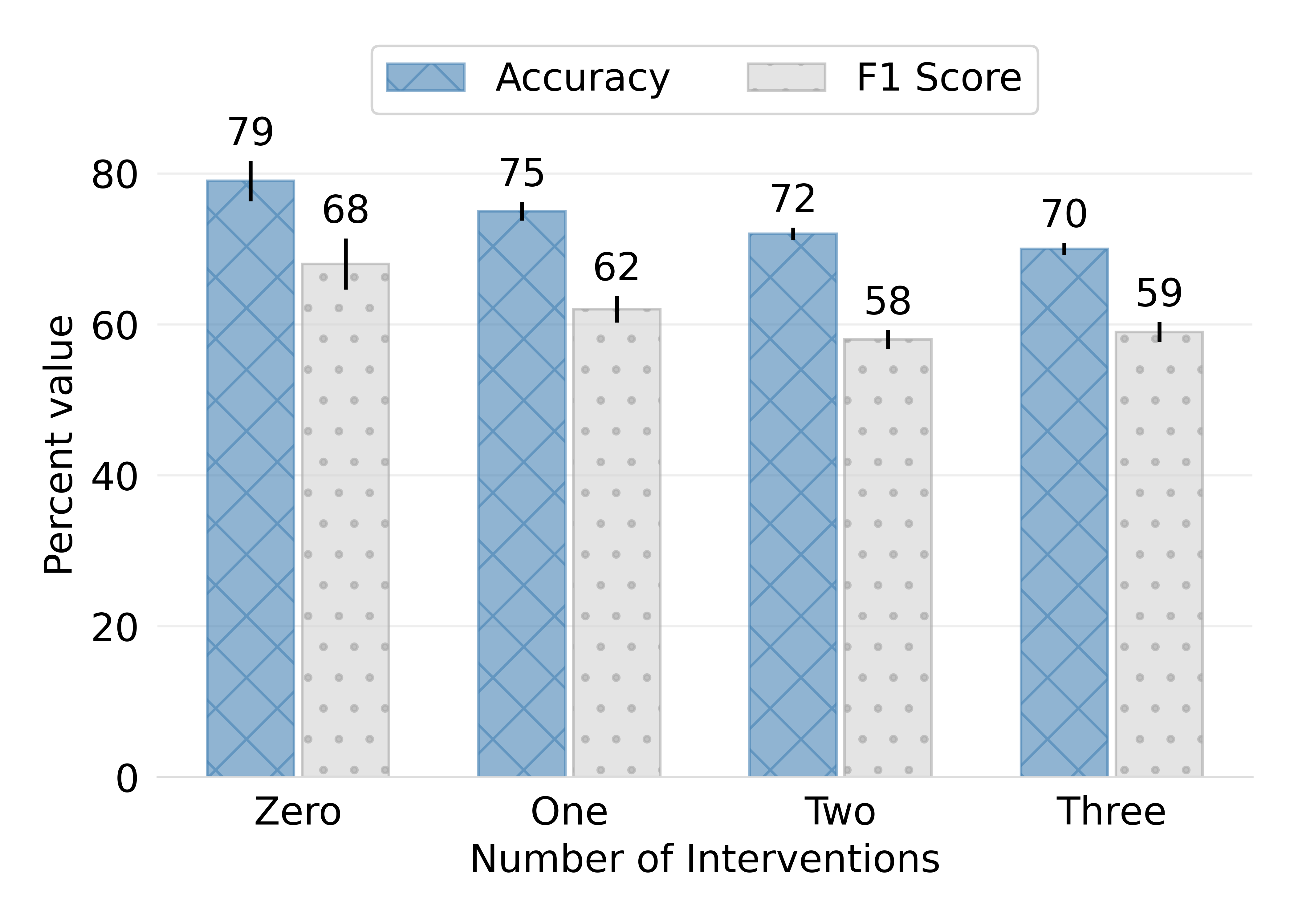}
        \setlength{\belowcaptionskip}{-2pt}
        \setlength{\abovecaptionskip}{-1pt}
        \caption{Mean Accuracy and F1 Score for different number of interventions. Error bars represent standard error. 
        %We observe a steady decrease in both metrics with more number of interventions.
        }  
        \label{fig:utility}
    \end{minipage}%
    \hspace{0.01\textwidth}
    \begin{minipage}{0.65\textwidth}
        \centering
        \includegraphics[width=0.9\textwidth]{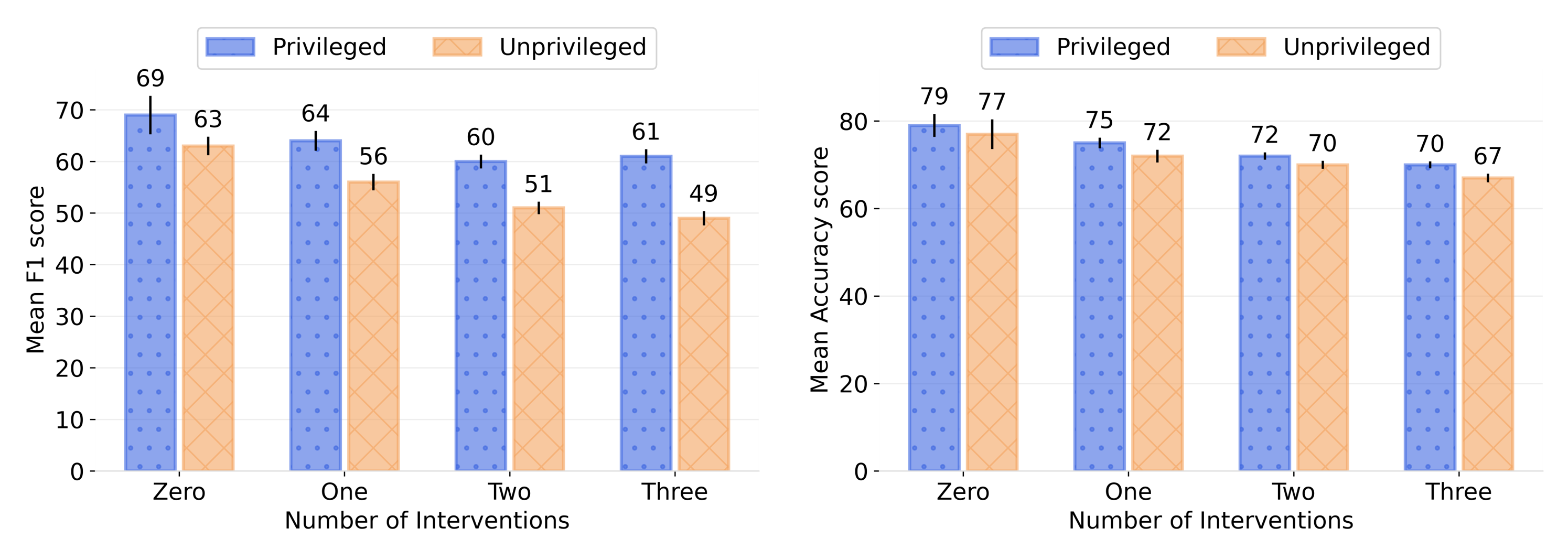}
        \setlength{\belowcaptionskip}{-2pt}
        \setlength{\abovecaptionskip}{-1pt}
        \caption{Mean F1 Score and Accuracy for the privileged and the unprivileged group across different number of interventions. 
        %We observe that both metrics decrease for the privileged and the unprivileged groups with more number of interventions. 
        Error bars represent standard error.}
        \label{fig:f1_score}
    \end{minipage}
\vspace{-1em}
\end{figure*}

It is important to note that all of these patterns reflect the aggregate trend and may not apply for all cases. For example, \autoref{fig:counter_exp} shows the absolute values for the accuracy difference metric across different interventions for the Adult Income dataset. In this case, we observe multiple instances where a larger number of interventions did not lead to more fairness (lower values). 
%On the other hand, we observed multiple instances where lower number of interventions performed better than higher numbers of interventions. 
This observation is contrary to the aggregate trend for group fairness that we observed in \autoref{fig:group_indi_inter}(A). So, \textit{it is not always the case that more interventions will result in more fairness. One needs to choose the right combination of interventions to get the best results.} We will discuss which combinations work for different metrics in \autoref{sec:table}.  %\textcolor{blue}{this figure does not show a single combination where more interventions benefited bias. None of the two-intervention bars is lower than the 1-intervention bars and the same goes for the 3-intervention bars. There is a clear upward trend. I assume smaller SPD numbers are better. I also worry that we might get accused of cherry picking since we just pick one of many metrics to show an outcome that helps our message.}
%For the second part, we present a few pictures of counter examples where bias scores increase compared to baseline. Try to include examples from different datasets and metrics. Here, each picture will be a bar chart corresponding to a bias metric and a specific dataset. The x-axis will show different interventions. 
%\autoref{fig:counter_exp} represents absolute percentage value for Statistical parity difference across different interventions for the Adult Income dataset. Here, lower values are desirable as they mean lesser disparity. We have selected an example which shows the more interventions does not always lead to more fairness. 

%\subsection{Effect of Cascaded Interventions on Utility Metrics}
\subsection{Effect on Utility Metrics (R2)}

We start off by analyzing how different number of interventions compare against each other on utility metrics as a whole. Following a similar procedure as defined in \autoref{sec:res_fair}, we plot a heatmap for utility metrics instead of fairness metrics (see \autoref{fig:heatmap} (B)). Here, a cell (i,j) represents the percentage of cases where j interventions yielded lower utility than i interventions. As expected, we observe that any non-zero number of interventions results in lower utility than the baseline case (see row i=0). Similarly, we observe that two interventions yields worse utility metrics than one intervention (see cell(1,2)) and three interventions yields worse utility metrics than two interventions (see cell(2,3)). Overall, this reveals a strong downward trend for utility metrics with more number of interventions. Looking at \autoref{fig:heatmap} (A) and (B) in conjunction, we observe that three interventions perform on par with two interventions on fairness. However, 75\% of the times three interventions performed worse on utility metrics than two interventions. This observation hints that one should typically opt for two interventions and go for the third intervention only in specific contexts.

To quantitatively understand the effect on specific utility metrics, %we analyzed how Accuracy and F1 score vary across different number of interventions. 
%We grouped all interventions across datasets into 4 buckets (0 intervention, 1 intervention, 2 interventions and 3 interventions) and then computed the mean accuracy and F1 score for each bucket. 
we computed the mean accuracy and F1 score for different number of interventions across datasets.
These mean scores are visualized in \autoref{fig:utility}. In line with our findings in \autoref{fig:heatmap}(B), we observe that both accuracy and F1 score steadily decrease as the number of interventions increase. 
%%%This downward trend is more pronounced in the beginning than the end. For e.g., the mean F1 score drops by 5\% going from no intervention to one intervention and later stabilizes going from two interventions to three interventions. Overall, 
This trend shows that there is a cost to be paid for adding more interventions. So, \textit{one should not blindly opt for more interventions.} ML practitioners should consider the potential loss in utility metrics while designing fair ML pipelines. 

\begin{table*}[]
\caption{Spearman correlation coefficient between F1 score and fairness metrics for different number of interventions (represented as rows). We observe that the correlation coefficient decreases as the number of interventions increase across metrics. }
\begin{tabular}{rrrrrrrrrr}
\hline
 & FPR Diff & FNR Diff & Accuracy Diff & FOR Diff & FDR Diff & SPD    & F1 Score Diff & Theil Index & Consistency \\
\hline
0                & 0.748    & 0.42     & 0.385         & 0.42     & 0.329    & 0.678  & 0.58          & 0.708       & -0.986      \\
1                & 0.343    & -0.054   & 0.211         & 0.354    & 0.176    & 0.263  & 0.253         & 0.148       & -0.628      \\
2                & 0.228    & -0.11    & 0.15          & 0.24     & 0.148    & 0.109  & 0.115         & -0.144      & -0.481      \\
3                & 0.119    & -0.185   & 0.189         & 0.058    & -0.112   & -0.056 & 0.073         & -0.282      & -0.167 \\ \hline    
\end{tabular}
\label{table:corr}
\end{table*}

\begin{figure*}
 \centering 
 \includegraphics[width=1.9\columnwidth]{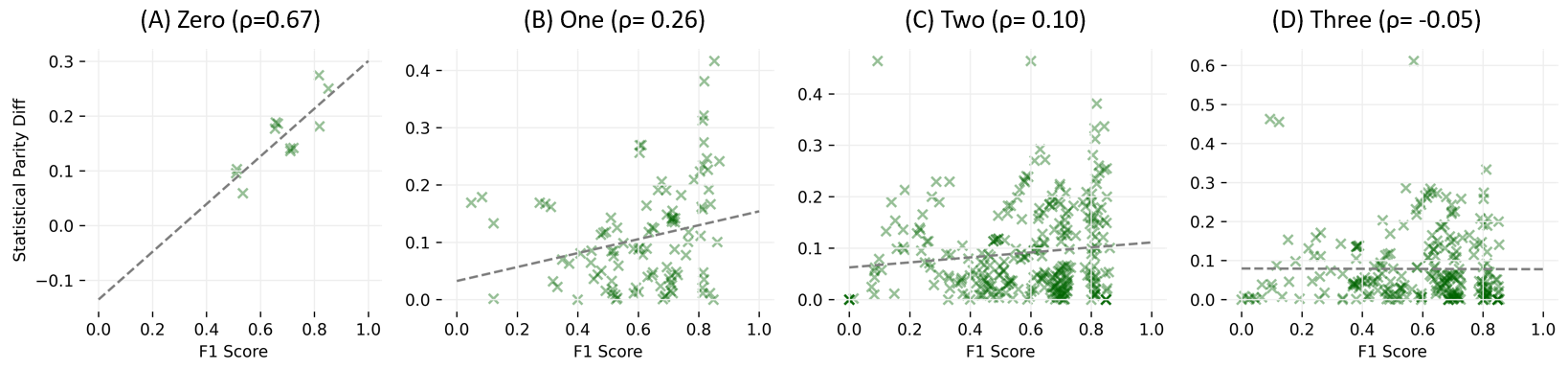}
 \caption{Relation between Statistical Parity difference and F1 score for different numbers of interventions from 0 to 3 (A - D). Each green `x' marker corresponds to a specific intervention executed on one of the 3 random subsets of a given dataset. The grey line represents the regression line that best fits all the points. It visually indicates the strength of the correlation.  }  
 \label{fig:corr}
\end{figure*}

So far, we have looked at the effect of cascaded interventions on utility metrics and fairness metrics in isolation. Now, let us investigate the effect of cascaded interventions on the bivariate relationship between utility metrics and fairness metrics. We start off by grouping all experimental data across datasets by the number of interventions. For each group, we compute the Spearman correlation coefficient between different fairness metrics and F1 score as shown in \autoref{table:corr}. Here, we have used absolute values for group fairness metrics. We observe that there is a significant positive correlation between fairness metrics and F1 score for the baseline case. For all fairness metrics except for the consistency metric, a higher value means more bias (less fairness). Hence, a positive correlation suggests that F1 score and fairness are negatively linked. In other words, \textit{interventions with high F1 score generally result in poor fairness and vice versa}. This observation is in line with existing literature which discusses a tradeoff between accuracy and fairness for individual interventions \cite{menon2018cost}. As the number of interventions increases, we observe a steady decline in the correlation coefficient across fairness metrics. Here, the correlation coefficient for consistency moves in the opposite direction as unlike all other fairness metrics higher values means more fairness. \textit{The decrease in correlation suggests that the likelihood of attaining a high F1 score along with high fairness increases with higher numbers of interventions}. As an example, we plot the bivariate relation between Statistical parity difference and F1 score for different numbers of interventions (see \autoref{fig:corr}). 
%We observe that three interventions are able to achieve high F1 score and low bias scores more consistently than one or two interventions. Here, 
The decrease in correlation coefficient $\rho$ is evident from the decrease in the slope of the regression line as we move towards higher numbers of interventions. If the reduction in F1 score caused by different interventions was in proportion to the corresponding increase in fairness, the correlation coefficient would have remained constant across different number of interventions. So, the observed decrease in correlation coefficient suggests the efficacy of cascaded interventions in reducing bias without sacrificing too much on performance (F1 Score). %\textcolor{blue}{this is also unclear. Fig 6 showed that SPD went up with more interventions but here it's going down.}  

%Looking at the effect of multiple interventions on fairness and utility in conjunction, we observe a tradeoff between utility and fairness.  Our work concurs with this line of work and takes it a step further by showing its presence for multiple interventions as well.  

%\subsection{Impact of intervention techniques on privileged group?}
%\vspace{-1em}
\subsection{Effect on Population Groups (R3)}

 In our experimental setup, we  kept a log of different statistics like false positive rate, false negative rate, F1 score, base rate, 
 %\textcolor{blue}{not sure if base rate matters. I read that base rate is just the number of positives over the entire data. It's not dependent on the classifier.}
 etc. for the privileged and unprivileged groups across all interventions. We analyzed this data to understand the impact of different interventions on these two groups.
 %compared to the baseline (no intervention). 
 \autoref{fig:f1_score} shows the aggregate impact of different number of interventions on accuracy and F1 score. In line with our earlier finding (see \autoref{fig:utility}), we observe that these utility metrics deteriorate for both groups with more number of interventions. However, the impact on the underprivileged group is more severe than the privileged group for the F1 metric. The F1 score for the privileged group dropped 8 percent points from 69\% to 61\% while it dropped 14 percent points for the underprivileged group from 63\% to 49\%. This disproportionate impact also lead to an increase in disparity between the groups in terms of F1 Score from 6\% to 12\%. 
 %Moreover, the disparity between groups also increases steadily from 6\% for the no intervention case to 13\% for the three interventions case. 
 In the case of accuracy, the impact on both groups is roughly even and the disparity between groups remains almost constant ($\scriptstyle\mathtt{\sim}$3\%) across different number of interventions.

\begin{figure*}
    \centering
    \begin{minipage}{.59\textwidth}
        \centering
        \includegraphics[width=0.9\columnwidth]{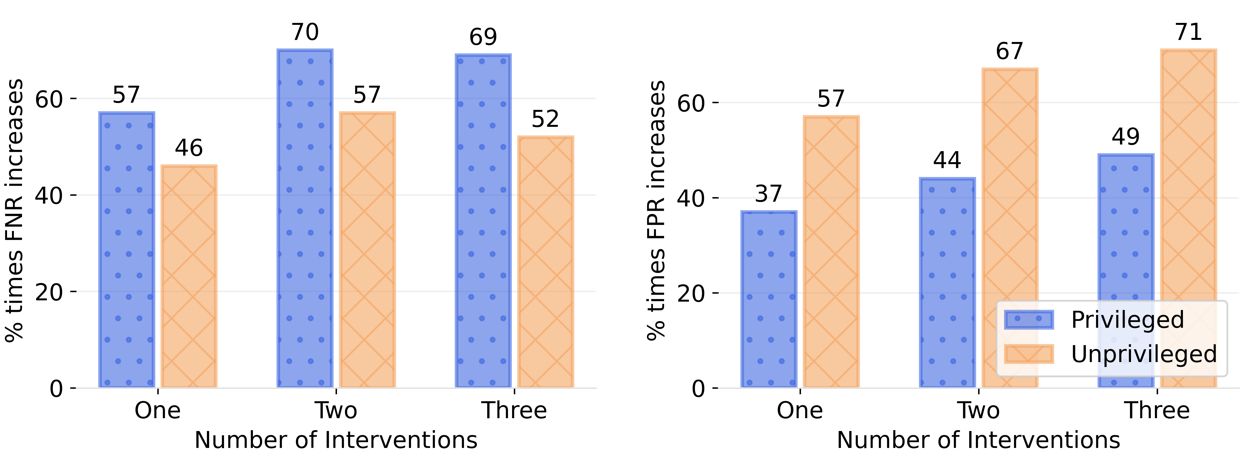}
        \caption{Percentage of times false negative rate (Left) and false positive rate (Right) increases compared to the baseline (No Intervention) across different number of interventions for all datasets.}  
        \label{fig:group_metrics}
    \end{minipage}%
    \hspace{0.01\textwidth}
    \begin{minipage}{0.36\textwidth}
        \centering
        \includegraphics[width=0.9\columnwidth]{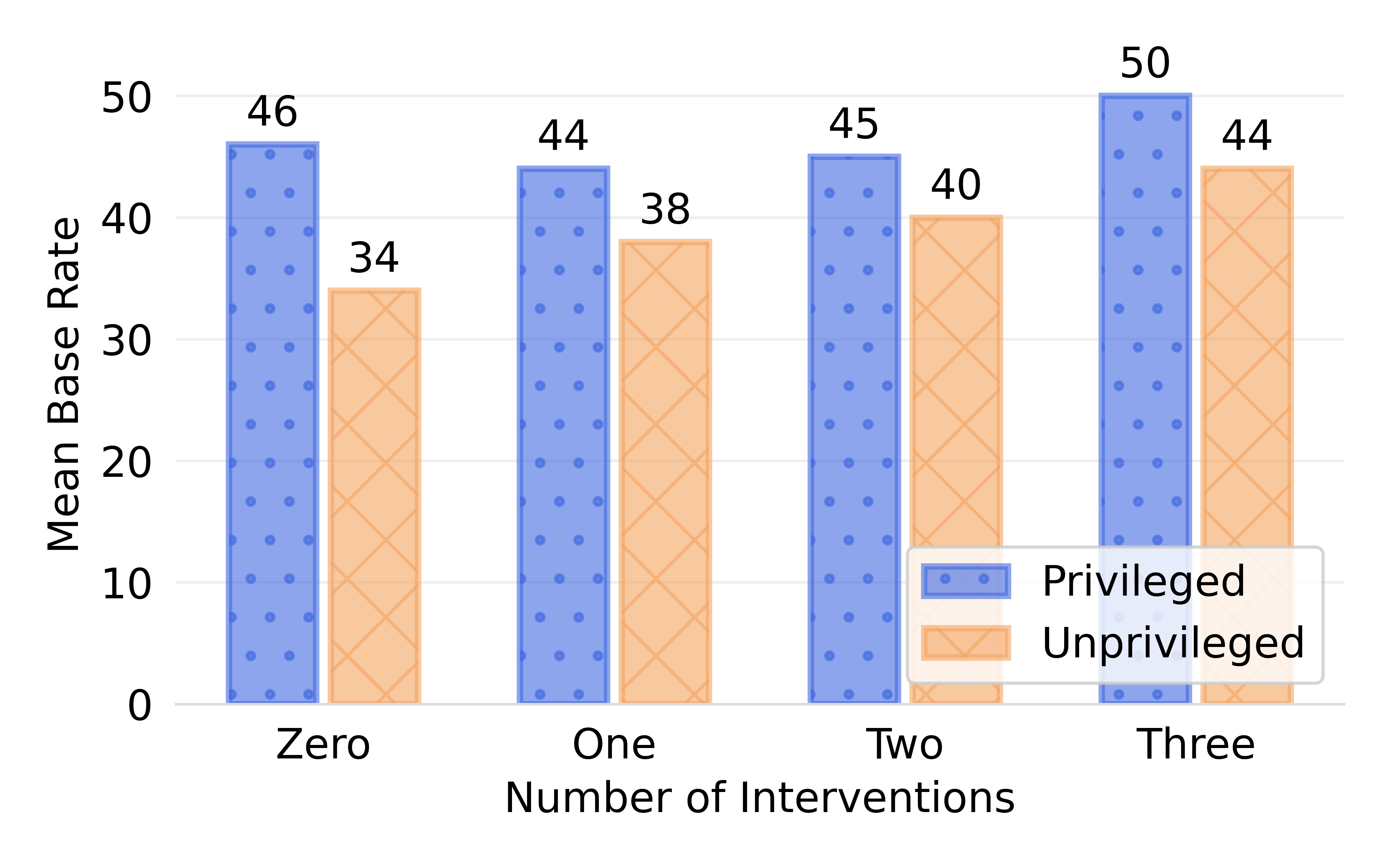}
        \setlength{\belowcaptionskip}{-2pt}
        \setlength{\abovecaptionskip}{-1pt}
        \caption{Mean base rate for the privileged and unprivileged groups across different number of interventions. 
        %We observe that the difference in base rate decreases with higher numbers of interventions.
        }
        \label{fig:base_rate}
    \end{minipage}
\end{figure*}

 The decrease in utility metrics signal an increase in error rates. So, let us look at the impact on the false positive rate (FPR) and false negative rate (FNR).   
 %More specifically, we compared the outcomes of different interventions with the baseline case and measured the number of times we got worse metrics than we started out with. 
 As shown in \autoref{fig:group_metrics},
 %shows the percentage of times we observed an increase in FNR and FPR compared to the baseline for different number of interventions. As 
 we observe a large percentage of cases where individual interventions resulted in higher error rates for both the privileged and the unprivileged group compare to the baseline (no intervention case). As we go for higher numbers of interventions, the percentage of such cases generally increases further. This trend is in agreement with the decreasing trend in utility metrics for more number of interventions. On comparing between groups, we find that interventions are more likely to result in higher FNR for the privileged group than the unprivileged one. It means that individuals from the privileged group are more likely to be misclassified with the unfavorable outcome than the unprivileged group. 
 %\textcolor{blue}{as it comes to FNR and FPR, when looking at the datasets you used in all cases FNR seems to be a bad thing for the person, like they get denied for something, but for the COMPAS data they don't get put in jail. There FPR is a bad thing, or unfavorable outcome as you call it. So I am not sure if that impacts the semantics of the message}  
 This trend flips for FPR where the unprivileged group are more likely to have a higher FPR. In other words, individuals from the unprivileged group are more likely to be misclassified with the favorable outcome than the privileged group. Both these trends 
 %persist for different numbers of interventions and generally 
 appear to deepen with higher number of interventions. %From these trends, it appears that different interventions positively impact the unprivileged group and negatively impact the privileged group. 
 Looking at these patterns in conjunction with \autoref{fig:f1_score}, it appears that the loss in Accuracy/F1 score can atleast be partially explained by the tendency of the interventions to assign more positive outcomes to the unprivileged group and negative outcomes to the privileged group.  
 %These trends show that different interventions are disproportionately assigning negative outcomes to the privileged group and positive outcomes to the unprivileged group.
 %Looking at both these trends together hints that different interventions are in effect transferring some of the positive outcomes from the privileged to the unprivileged group.  
 %We will investigate this pattern further when we look at the base rate. 
 %However, this holds true under the assumption that the total number of positive outcomes is roughly constant. 
 %This also explains the reduction in disparity between base rates.
 
 %We investigated this pattern further by analyzing how the base rate vary for different groups with more number of interventions (see \autoref{fig:base_rate}).  
 Next, let us look at the impact on base rate for different groups (see \autoref{fig:base_rate}). 
 Here, base rate is defined as the proportion of positive outcomes for different groups. It is computed over model's prediction for the test data post all relevant interventions. For the no intervention case, we observe a 12\% disparity in favor of the privileged group. With more interventions, the base rate for the unprivileged group steadily increases from 34\% to 44\% (10 \% jump). On the other hand, the base rate for the privileged group decreases a bit for the one and two interventions case and increases by 4\% for the three interventions case. 
 %This observation concurs with our earlier finding where we observed disproportionate increase in FPR for the unprivileged group. 
 Overall, this leads to a decrease in disparity between groups from 12\% to 5\% for the two intervention case. In a context where equality between base rates is a priority, two interventions seems to be the way to go. It is also important to note that some of the interventions can negatively impact the privileged group. This is evident from the drop in base rate for the privileged group for the one and two interventions case. 
 %For the one and two interventions case, the base rate for the privileged group decreases and the unprivileged group incr
 If we look at base rate over the entire population, we find that the base rate undergoes a modest increment for the one (1\%) and two interventions case (2\%). However, it increases significantly for the three interventions case (7\%). So, ML practitioners should exercise caution while adding the third intervention, especially for contexts where the number of favorable outcomes is fixed such as hiring.  
 %\textcolor{blue}{this the trends of this figure seem to point out something important. It should be brought out more persuasively but since it uses base rate I am not sure what it is.} 
 %On analysis of this data, we found multiple cases where disparity was reduced by making things worst for individual groups.     
 
\begin{figure*}
 \centering 
 \includegraphics[width=1.7\columnwidth]{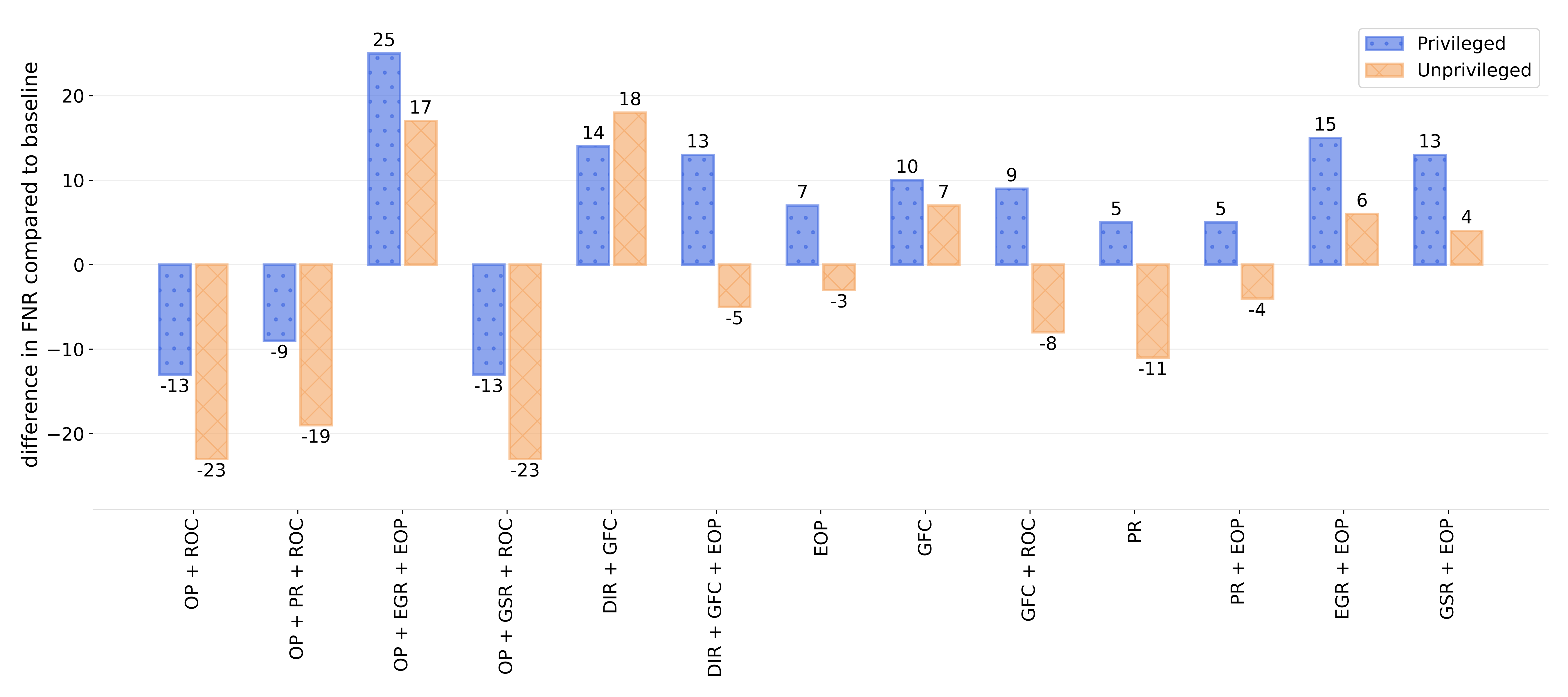}
 \caption{Change in false negative rate compared to the baseline across different interventions for the Adult Income dataset. Here, negative values are desirable.}  
 \label{fig:fnr_adult}
\end{figure*}

%There seems to be some discrepancy when we look at \autoref{fig:group_indi_inter} and \autoref{fig:group_metrics} in conjunction. \autoref{fig:group_indi_inter} shows that group fairness metrics decrease with more number of interventions. However, we do not observe a similar reduction in error rates for individual groups as shown in \autoref{fig:group_metrics}. 
Fairness metrics provide elegant mathematical representations for different notions of fairness but might obscure the specific effect on different population groups. \autoref{fig:group_indi_inter} shows that group fairness improves with more interventions on aggregate. However, we do not observe a reduction in error rates for individual groups as shown in \autoref{fig:group_metrics}. To investigate this further, let us look at a specific example pertaining to the Adult Income dataset. \autoref{fig:fnr_adult} shows the difference in false negative rates (FNR) compared to the baseline for the privileged (males) and the unprivileged (females) groups across different interventions. Here, we have only considered interventions that have reduced the magnitude of the FNR difference fairness metric compared to the baseline. So, as per the FNR Diff metric, all of these interventions are effective at reducing bias. However, we observe that for 10 out of the 13 interventions FNR has actually increased for one or both groups (see \autoref{fig:fnr_adult}). 
%In , we observe 10 cases (interventions) where FNR has \textit{increased} for either or both groups. 
So, the reduction in the FNR Diff metric is partly due to an \textit{increase} in FNR for either/both population groups.
%It can be argued if it is desirable to reduce disparity by increasing error rates unevenly for different groups. 
Such means (interventions) of reducing disparity can be deemed as socially undesirable. 
Existing group fairness metrics solely focus on disparity and are indifferent to how disparity is reduced. \textit{This finding points to a need for new fairness metrics that account for the specific impact on individual groups apart from just the gap between those groups}. Such metrics should give preference to interventions that reduce disparity by improving error rates for all population groups (at least not making it worse for any population group). 

%As we are dealing with Adult Income dataset, false negatives represents high income individuals who were misclassified as low income. A positive change in false negative rate means that the proportion of such misclassifications has increased and vice versa.   Ideally, we would not likemake things better for all groups. 

There can be different ways to interpret the empirical findings based on one's value system. One way to interpret these numbers can be from a pure ML perspective where the 
%dataset is sacrosanct as it captures reality. 
focus is to train a ML model that best fits the underlying dataset.
Here, the objective of an intervention is to ensure that the model performs equally well for different groups in terms of accuracy, false positive rate, false negatives rate, etc. 
%In the context of AI fairness, this will mean that the interventions should aim to reduce errors rates so that the ML model performs equally well for different groups  
From this viewpoint, many of the interventions are counterproductive as they increase error rates and decrease accuracy/F1 score for different groups (see \autoref{fig:f1_score} and \autoref{fig:group_metrics}). Going from individual interventions to cascaded interventions makes things worse as the error rates for population groups increases further and the accuracy/F1 score further deteriorates. Ideally, one would reduce disparity by reducing the error rates for different groups but not making it worse for any of them. For example, 3 interventions in \autoref{fig:fnr_adult} reduce disparity (FNR Diff metric) by reducing FNR for both groups. 
%, namely OP + ROC, OP + PR + ROC and  OP + GSR + ROC. 
Similarly, mitigating discrimination against one group at the cost of another is self-defeating and unethical.
%and might stifle research efforts in this direction. 
For illustration, we find 5 interventions in \autoref{fig:fnr_adult} where FNR decreases for the unprivileged group and increases for the privileged group compared to the baseline. In other words, some high income individuals from the privileged group were misclassified as low income in an effort to increase fairness. This is not a one off case. Our experiments show an aggregate trend across datasets where privileged groups were disproportionately misclassified with unfavorable outcomes and unprivileged groups were disproportionately misclassified with favorable outcomes (see \autoref{fig:group_metrics}). These observations highlight how interventions can negatively impact the privileged group.
%, and the finding is further supported by the disproportionate loss in F1 score across datasets for the privileged group (see \autoref{fig:group_metrics}). 
Future interventions should be more considerate towards their possible negative impact on the different population groups.

%\textit{As reflected from our experiments, fairness enhancing interventions can negatively impact the privileged group. For certain instances, we found that fairness enhancing interventions can cause discrimination against the privileged group.} This is a critical finding which can significantly impact how we see existing interventions and how future interventions should be devised. Mitigating discrimination for one group at the cost of another is self-defeating, unethical and might stifle research efforts in this direction. 
Another way to interpret these numbers can be from the perspective of social justice where interventions should not only reduce disparity in error rates but also serve as an instrument to right historical wrongs. For example, interventions should enforce equality/equity of outcomes, i.e., ML models should assign positive outcomes to different groups at the same rate or even prioritize the unprivileged group irrespective of the patterns in the dataset \cite{mehrabi2020statistical}. In the process of achieving these goals, the loss incurred in terms of accuracy and F1 score is secondary as these metrics are computed over output labels (ground truth) polluted with historical biases. Similarly, the disproportionate increase in FNR for the privileged group is incidental/imperative to serve the larger goal of equal representation. Under this viewpoint, 
%existing interventions are doing a decent job in supporting the unprivileged group get to an equal footing. 
cascaded interventions are desirable as they help bridge the gap in base rates (see \autoref{fig:base_rate}).  

%People may have different value systems and so it is difficult to provide a general guideline. 
%We have presented empirical evidence on the impact on interventions on the privileged/unprivileged group and different ways to interpret it. 
%Based on one's value system, they can account for these findings while designing fair ML pipelines. 
%Irrespective of one's value system, we believe in making things transparent for people to see fairness issues along different perspectives. \textcolor{blue}{incomplete sentence, too colloquial, rephrase or drop} 

%Fairness literature has primarily focused on the impact of algorithms on underprivileged groups. Here, we would like to highlight the impact of individual interventions as well as cascaded interventions on the privileged groups as well.    
%Future research on fairness enhancing interventions should make sure that they don't have a negative effect on the privileged group. For example, reducing False Negative rate difference should be achieved by reducing the error for the unprivileged group instead of increasing error for the privileged group. 

\begin{table*}[]
\caption{Ranking of 10 best and worst performing interventions for different evaluation metrics. Here, we have ranked interventions in ascending order of their corresponding absolute values for all metrics except for Accuracy, F1 Score and Consistency that are sorted in descending order. This ordering schema ensures that desirable interventions are ranked higher for all metrics. Interventions with `+' sign represents combinations of multiple interventions. Here, `Logistic Regression' represents the baseline case, i.e., no intervention for all stages.  }

\begin{tabular}{llllll}
\hline
Rank & Accuracy            & F1 Score            & Theil Index         & Consistency     & FPR Diff            \\
\hline
1    & Logistic Regression & DIR + EGR + ROC     & DIR + EGR + ROC     & OP + GFC + ROC  & EGR + EOP           \\
2    & DIR                 & DIR                 & DIR                 & OP + GFC + CEOP & OP + GFC + ROC      \\
3    & GSR                 & DIR + GSR           & DIR + EGR           & OP + GFC        & PR + EOP            \\
4    & DIR + EGR + ROC     & DIR + EGR           & DIR + GSR           & OP + GSR        & OP + EGR + EOP      \\
5    & DIR + EGR           & Logistic Regression & DIR + PR + EOP      & OP              & OP + GFC + EOP      \\
6    & DIR + GSR           & OP + CEOP           & DIR + ROC           & OP + EGR + ROC  & OP + GFC            \\
7    & EGR                 & OP + GSR + CEOP     & DIR + CEOP          & OP + PR         & DIR + GFC           \\
8    & DIR + PR + CEOP     & OP                  & Logistic Regression & OP + GSR + CEOP & GFC + EOP           \\
9    & CEOP                & OP + GSR            & DIR + PR + ROC      & OP + CEOP       & DIR + EOP           \\
10   & PR + CEOP           & GSR                 & DIR + GSR + CEOP    & OP + PR + CEOP  & DIR + GSR + EOP     \\
&&&&& \\
51   & OP + GFC + CEOP     & OP + PR             & OP + PR + ROC       & PR + EOP        & Logistic Regression \\
52   & OP + PR + ROC       & OP + GFC + EOP      & OP + PR             & DIR + PR + ROC  & GSR + CEOP          \\
53   & DIR + GFC + EOP     & DIR + GFC + EOP     & OP + ROC            & DIR + EGR + EOP & OP                  \\
54   & DIR + GFC           & GFC + EOP           & OP + GFC            & DIR + EOP       & PR + CEOP           \\
55   & OP + ROC            & OP + GFC            & DIR + GFC           & DIR + GFC + EOP & GFC + CEOP          \\
56   & OP + GFC + EOP      & DIR + GFC           & GFC + CEOP          & EOP             & DIR + GSR + CEOP    \\
57   & OP + GSR + ROC      & GFC + ROC           & OP + GFC + EOP      & OP + GSR + EOP  & DIR + PR + CEOP     \\
58   & DIR + PR + ROC      & DIR + GFC + CEOP    & GFC + ROC           & OP + EOP        & OP + CEOP           \\
59   & DIR + ROC           & GFC + CEOP          & OP + GSR + ROC      & OP + PR + EOP   & DIR + CEOP          \\
60   & OP + PR             & DIR + GFC + ROC     & DIR + GFC + ROC     & OP + GFC + EOP  & CEOP        \\ \hline        \\
\end{tabular}
%\vspace{-2em}
\begin{tabular}{llllll}
\\  \hline
Rank & FNR Diff         & Accuracy Diff   & FOR Diff         & FDR Diff            & SPD \\
\hline  
1    & OP + GFC + EOP   & DIR + GFC + ROC & OP + PR          & Logistic Regression & OP + GFC + EOP          \\
2    & GFC + EOP        & DIR + GFC       & DIR + GFC + CEOP & DIR                 & OP + EGR + ROC          \\
3    & OP + EGR + EOP   & OP + PR + ROC   & CEOP             & GSR + CEOP          & OP + EGR                \\
4    & OP + EOP         & OP + ROC        & PR + CEOP        & DIR + EGR + CEOP    & OP + GSR + EOP          \\
5    & OP + GSR + ROC   & ROC             & DIR + CEOP       & GFC + CEOP          & OP + GFC + ROC          \\
6    & OP + GFC + ROC   & EOP             & PR               & GFC                 & GSR + ROC               \\
7    & OP + PR + EOP    & DIR + GFC + EOP & DIR + PR + CEOP  & DIR + GFC           & DIR + GFC               \\
8    & DIR + GSR + EOP  & OP + GSR + ROC  & OP + GFC + ROC   & DIR + PR            & OP + EOP                \\
9    & GSR + EOP        & GFC + CEOP      & DIR + GSR + CEOP & DIR + EGR           & EGR + ROC               \\
10   & OP + GFC         & PR + ROC        & GFC + CEOP       & PR                  & OP + PR + EOP           \\
&&&&&  \\
51   & GFC + CEOP       & OP + EGR        & GSR + ROC        & OP + GFC + CEOP     & GFC + CEOP              \\
52   & DIR + EGR + CEOP & OP + PR         & GFC + EOP        & DIR + GSR + EOP     & PR                      \\
53   & GSR + CEOP       & OP + PR + CEOP  & OP + ROC         & OP + EGR            & OP + CEOP               \\
54   & DIR + GFC + CEOP & OP + EGR + ROC  & OP + CEOP        & OP + EGR + ROC      & GSR + CEOP              \\
55   & EGR + CEOP       & OP + GSR        & OP + GSR + ROC   & OP + PR             & DIR + PR + CEOP         \\
56   & DIR + GSR + CEOP & OP + GSR + CEOP & OP + PR + ROC    & OP + EOP            & Logistic Regression     \\
57   & DIR + PR + CEOP  & OP + GFC + ROC  & OP + EGR + EOP   & OP + EGR + EOP      & DIR + GSR + CEOP        \\
58   & PR + CEOP        & OP + EGR + CEOP & DIR + GFC + EOP  & OP + PR + EOP       & PR + CEOP               \\
59   & DIR + CEOP       & OP + GFC        & OP + EGR + ROC   & OP + GFC + EOP      & DIR + CEOP              \\
60   & CEOP             & OP + GFC + CEOP & OP + EGR         & OP + GSR + EOP      & CEOP                    \\ 
\hline
\end{tabular}

\label{table:rq4}
\end{table*}

%\vspace{-1em}
\subsection{Comparison between Interventions (R4)}
\label{sec:table}
%In the previous sections, 
So far, we have focused on the aggregate trends which might or might not apply for a given intervention. In this section, we focus on specific interventions and how they compare against different evaluation metrics. 
%Such knowledge can assist practitioners and researchers in choosing interventions based on their specific context. 
%%%We begin by computing the mean scores for all evaluation metrics across 4 datasets corresponding to each intervention.
%our investigation by grouping all experimental data based on different interventions and then computed the mean across the 4 datasets. 
%%%This provided a mean score for each evaluation metric across 60 different interventions. 
For each evaluation metric, we ranked all interventions from the best performing to the worst performing based on their respective mean score across datasets. For the accuracy, F1 score and consistency metric, higher values are desirable so we sorted interventions based on descending order of their corresponding values. For all other metrics, we used ascending order. Moreover, we used absolute values for all group fairness metrics as we are primarily concerned with the magnitude of bias.
%and not which group is discriminated against.    
%For each of the 4 datasets, we conducted 60 different interventions. This process was repeated 3 times with different train test splits.      

\autoref{table:rq4} contains the 10 best and worst performing interventions for different evaluation metrics. From this table, we can make a few important observations. Logistic regression (by which we mean `No intervention') tops the list for the accuracy metric. 
%This is in line with existing literature that comments on the accuracy-fairness tradeoff \cite{berk2017convex}. 
As all interventions optimize for some aspect of fairness, they might sacrifice a bit on accuracy. So, one should not use any intervention for achieving the best accuracy.  For the F1 score, we observe that a few interventions rank higher than Logistic Regression, such as DIR + EGR + ROC. However, the difference between them was quite slim (0.3\%) which might be attributed to imbalanced output class distribution. The broader point is that \textit{applying more interventions does not always lead to a significant loss in utility}. In the case of DIR + EGR + ROC, we observe the best performance for the F1 score and the 4th best for accuracy. Among the 10 bottom ranked interventions across all fairness metrics, Logistic Regression occurs only twice. This shows that there are several individual and cascaded interventions that perform worse than the baseline case for at least some fairness metric. Hence, it is important to choose interventions wisely. 
%We hope empirical studies like ours will assist in doing just that. 
ML practitioners/researchers can leverage resources like \autoref{table:rq4} and prioritize interventions that have worked well for other datasets while designing their own fair ML pipeline(s).

For the fairness metrics, we observe that the best performing intervention is mostly unique for each one of them. In other words, there is no silver bullet for all fairness metrics. 
%e intervention that which outperforms all other interventions for all or even a majority of evaluation metrics. 
This observation is in line with the existing literature which proves that no intervention can simultaneously optimize for all fairness metrics \cite{kleinberg2016inherent}. From a practical standpoint, this implies that ML practitioners need to prioritize which metrics are more important to them and then choose interventions accordingly. It is also worth noting that the best performing intervention for any metric is either Logistic Regression (No Intervention) or a combination of two or more interventions. Apart from the top performing interventions, we also observe that the top 10 interventions for all fairness metrics are predominantly cascaded interventions. For example, 9 out of the top 10 interventions for the Consistency metric are cascaded interventions. These observations further motivate the efficacy of cascaded interventions over individual interventions. Among the top 10 interventions across all metrics, OP + GFC + ROC occurs the most number of times. Similarly, OP + GFC + EOP occurs the most number of times among the worst 10 interventions across metrics. It is interesting to see that both of these interventions have much in common (OP and GFC). This shows that certain intervention are more compatible/incompatible with another. Changing an ingredient can drastically impact the outcome. For instance, swapping ROC with EOP resulted in the entire combination (OP + GFC + ROC) to change from being one of the top ranked to one of the worst ranked interventions. 
%As expected, there is no single or combination of interventions that excel at all the evaluation metrics. Data Scientists/practitioners can pick and choose based on their respective context. 
It should be noted that ranking abstracts the real difference in magnitude. For brevity, we have used ranking in the table. 
We encourage the readers to refer to the source code/experimental data for more details. 

%\vspace{-0.5em}
\section{Discussion}
This work explores the realm of cascaded debiasing by asking basic research questions and then answering them via a large empirical study. The goal of this study is to understand the viability and possible fallouts of using cascaded fairness-enhancing interventions in terms of fairness, utility and impact on individual groups. To conduct such a study, we chose IBM's AIF 360 toolkit as it supports one of the largest collection of interventions, fairness metrics and datasets. On the flip side, we were limited to the different options it supported and faced runtime issues executing a certain intervention (Optimized Preprocessing)
%like the Optimized Preprocessing intervention for the Bank Marketing dataset 
for a particular dataset (Bank Marketing dataset). 
%We hope this package and other similar packages \cite{lee2021landscape} like fairlearn will keep evolving and resolve such issues in the future.
It is important to note that all insights and analyses presented in this paper are empirical in nature, and so, they may or may not generalize to other datasets, interventions or metrics. Having said that, our study covers a wide range of popular fairness metrics and state-of-the-art interventions. The insights reported in this paper can serve as a good starting point or heuristic to assist ML practitioners design fair ML pipelines and inform the design of fair ML tools. Moreover, these insights can help guide further research into this area. For e.g., why do certain combinations of interventions work better than others?  
%We hope the insights provided by this study will help guide further research, assist ML practitioners design fair ML pipelines, and inform the design of fair ML tools. 
%Since this is an empirical study, the insights and analyses presented in this paper may or may not generalize to other datasets, interventions or metrics. However, they can be a good starting point for ML 

%Despite the sizeable empirical study, our work only scratches the surface of the underlying problem and 
This work suggests multiple venues for future research. One research direction can be to conduct even larger studies which include more datasets \cite{ding2021retiring}, ML models, fairness metrics like counterfactual fairness \cite{kusner2017counterfactual}, statistical equity \cite{mehrabi2020statistical}, etc., other stages of intervention like the data curation stage \cite{ghai2020measuring}, and multiple hyperparameters for different interventions \cite{wu2021fair}. Such a study will paint a more comprehensive picture and its results will be more generalizable to different contexts. Another interesting research direction will be to conduct similar studies for other data types like text, images, etc., and consider other problem types such as regression, clustering, etc. This work deals with fairness at a group level (say males and females) and at an individual level (through the individual fairness metrics). It will be interesting to study the effect of cascaded interventions on different subgroups say black females, high-earning white males, etc. Future work might also take a deep dive into the underpinnings of the different trends/patterns reported in this paper. Lastly, the source code and experimental data can be accessed at \href{https://github.com/bhavyaghai/Cascaded-Debiasing}{\textit{github.com/bhavyaghai/Cascaded-Debiasing}} for easy reproducibility and for 
anyone to analyze/extend this study as they see fit. For e.g., one may choose to rerun this study using a different ML model, say SVM, or add other metrics/datasets.
\begin{acks}
This work was partially funded by NSF grants CNS 1900706, IIS 1527200, IIS 1941613, and NSF SBIR contract 1926949.
\end{acks}

\bibliographystyle{ACM-Reference-Format}
\bibliography{references}

\end{document}